# Knowledge Retrieval

Vishnu Vardhan Reddy Palli (Computer Science)

University of South Florida

*Abstract— Robots are man-made machines, which are used to accomplish the tasks. Robots are mainly used to do complex tasks and work in hazardous environment where humans are difficult to work. They are not only designed to use in hazardous environment but also in the environment where humans are performing the same task repeatedly. These are also used for cooking purpose; some tasks can be completed with the interaction of both the human and robot one of such things is cooking where human should help robot in making dishes. This paper mainly focusses on Functional Object-Oriented Network which is structured knowledge representation using the input, output and motion nodes. Task tress are generated using the task tree FOON is produced, and collections of all FOON's forms the universal FOON. Different algorithms to traverse the tree in order to get the best output are also discussed in this paper. The desired node or goal node can be achieved from the start node using the different search algorithms and comparison between them is discussed.*

## I. INTRODUCTION

The main aim of the robotics is to perform the tasks which are performed by humans daily, without intervention of humans. In order to perform these tasks autonomously robots require huge set of data that has been performed previously and what is the specific action that has been taken with the specific input. Robots also require knowledge representation along with several modules, to perform the action based up on the environment. Robots can understand human intentions and take actions accordingly. We have seen many robots helping humans in food delivery and cooking as well as serving the food.

Let's consider a robot making a food, first the environment in which humans work is dynamic in nature. In this environment only the robots should work. All the ingredients may or may not be available on the table. The robot should also recognize the ingredients by its structure because it cannot find the ingredient by its smell. In order to identify the ingredient by its state and gesture is highly complicated. The robot should be very much familiar with all the ingredients so that it can identify the ingredient correctly without any failure.

As our environment is dynamic in nature all the objects may not be available in the kitchen, it must be prepared by using the other objects which are available in the kitchen. Handling things in the kitchen is also very complicated. For example, egg must be handled very carefully else there is higher chances that egg might break. The robot must be familiar how to break the egg. Therefore, in order to achieve all these things to the fullest robot must be trained with huge data so that it can perform better.

Though we have all the ingredients in the kitchen and robot picks up any of the way possible to prepare a dish that might not be the optimal way to make a dish. In order to get the optimal way to make a dish we also use search algorithms that makes decision to select the node. Suppose consider we have a resource we should use the resource to the full extent we should not waste the resource, here we have robot we should use that in a precise way by not picking up the path which takes so long time, more resources i.e., kitchen items etc. Therefore, in order to optimize the searching mechanism, the search algorithms are used.

Based up on the type of algorithm the way of searching differs i.e., always selecting the first node, selecting node with minimal inputs, adding success rate to the actions that has to be performed, so that the robot can achieve the task in the optimal way and with minimal failure. By using these algorithms out of all the multiple paths available which lead to a similar output based up on the type of algorithm used the node will be selected.

Functional Object-Oriented Network (FOON) [1] is a knowledge representation for robots, describes the functional relationship between the objects and actions performed. With the interaction between the humans and robots we can achieve so many things. There are some situations where humans cannot afford any help like when a person is stuck in the fire, humans cannot get into the fire and rescue them, there are limited chances of these but when coming to robots, they can rescue the human life. But in some cases, robots cannot perform some tasks in such cases humans should support the robots in order o fulfill the tasks. Like in cooking, the human acts as an assistant to the robot, who has the expertise or knowledge to complete the task.

## II. Video annotation and FOON creation

Functional Object-Oriented Network (FOON) is a network which contain input nodes, output nodes and an action corresponding to the input and output nodes. Functional unit represents [2] actions in FOON by describing the state change of objects before and after execution, and it has input object nodes, output object nodes, and a motion node i.e., action taken.

The collection of input, output and motion node is called Functional unit. For example, let's consider breaking an egg as a functional unit, first we take an egg from basket which is input node and the motion node is breaking and the output node is broken egg.[5] Object nodes i.e., input and output nodes are denoted in green color and motion nodes are denoted as red color in FOON.

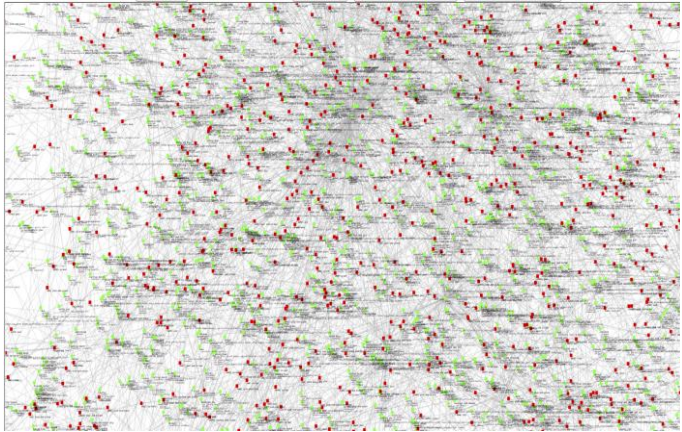

*Figure 1: Illustration of Universal FOON, which is collection of all sub graphs.*

FOON's are created using video annotations. Sub graph is termed as a FOON representing the single activity i.e., single recipe. Sub graph contains many functional units which describes the input, output nodes with all the different states of an object and the actions taken. As stated in figure 1 collection of all the subgraphs make Universal FOON.[4] Universal FOON will be very huge as the number of sub graphs increases

Task tree signifies the sequence of actions that need to be taken to achieve the goal. If we have a goal node, by using the task tree retrieval the robot along with the help of human will be able to achieve the optimal goal.[6] Based up on the experience we assign success rates to the actions, the action which can be done perfectly by the robot have the higher success rates, and the actions which robot is poor at is have the lower success rates. In the case of lower success rates, the humans should help the robots to perform the particular action.

```
//
O       bottle
S       contains{vanilla extract}
O       vanilla extract
S       in      [bottle]
O       spoon
S       empty
M       pour
O       spoon
S       contains{vanilla extract}
O       vanilla extract
S       in      [spoon]
//
```

*Figure 2: Illustration of single functional unit in text format, objects before the motion node are termed as input nodes and objects after motion node are termed as output nodes. Each functional units starts and ends with "//".*

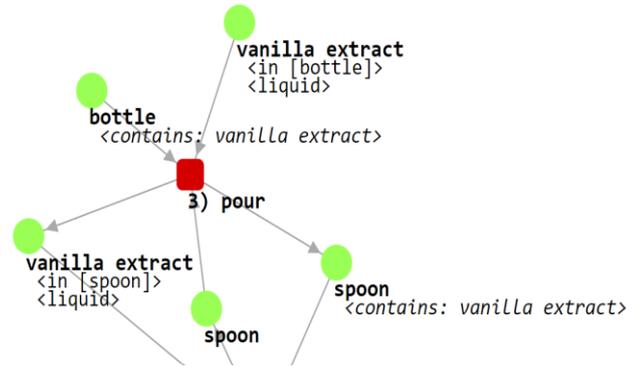

*Figure 3: Example of single functional unit with two input objects, one motion node 'pour' and two output objects.*

Figure 3 shows a functional unit with motion node as pour and input nodes are vanilla extract and bottle and output in vanilla extract in spoon. Figure 2 is the text representation of functional unit in figure 2. Before performing the action pour, vanilla extract is in the bottle and in liquid state, the input object bottle contains vanilla extract. The action pour is performed, pour is here treated as motion node. After the action is performed the output nodes or objects are spoon which contains vanilla extract and the vanilla extract is also the output node which is in spoon.

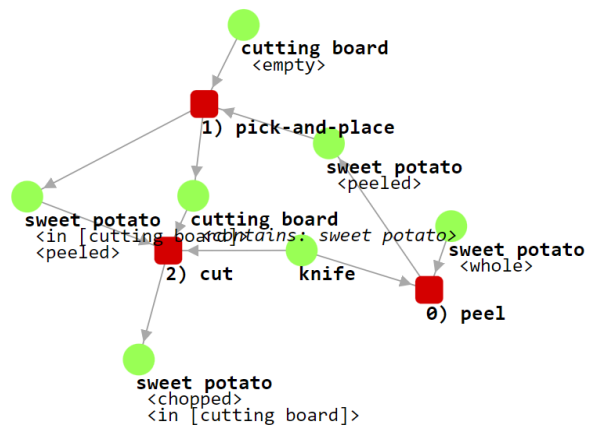

*Figure 4: This figure demonstrates the sub graph which is collection of three functional units.*

The figure 4 is sub graph, which is collection of functional units with different input and output nodes.

1) The input object is sweet potato, which is unpeeled, and the motion node is peel the action peel is performed using the knife, and the output by performing peel action is sweet potato in state of peeled.

2)The peeled sweet potato obtained from the previous state is the input for present functional unit along with the cutting board, which is empty currently, the motion node is pick and place that means we are picking up a sweet potato and placing it on the cutting board, the output node is cutting board which contains peeled potato.

3) For the last functional unit, the input node is cutting board with sweet potato along with the knife which is used for cutting and the motion node is cut, and the output node is sweet potato which is in chopped state on cutting board. These actions need to be performed in the sequence in order to get the desired results.

Task retrieval tree shows the sequence in which the actions need to be performed. In the above figure there are three functional units i.e., three motion nodes peel, pick and place and cut. The first functional unit contains two input nodes and one output node, for the next functional unit there are two input nodes and two output nodes and for the final functional unit there are three input nodes with one output node.

## III. METHODOLOGY

We have the sub graph which is the collection of all the functional units. The dish can be prepared in many ways based up on the input and output objects. While you explore all the nodes you find that there may be many nodes which leads to same object. Let's consider the case a particular output can be achieved by taking ten actions, but in the other case by taking only four actions the output can be achieved then we consider the second one as the optimized based up on some factors. Similarly, we have the search algorithms to be considered for FOON, in order to search for the goal node. The algorithms are BFS, DFS, Iterative deepening search and Greedy Best First Search.

The different files which are used in the project are: motion.txt contains all the different motions and their success rates, FOON.txt is the universal FOON which is combination of all the subgraphs, goal_nodes.json is the file which contains all the goal nodes we have to search for goal nodes using the algorithms, kitchen.json consists of all the items which are present in the kitchen.

a) Breadth First Search (BFS) is one of the algorithms in which the elements which are at the same level are searched first and then the next level and this continues until the goal node is found. This algorithm searches node in level-by-level fashion, this requires more space. As we are exploring all the nodes at the same level i.e., we are searching for the goal node at a particular level covering all the elements at that level, so it is known as BFS. BFS is one of the optimized algorithms to search for the goal node. The problem with this approach is if the node is present at depth 'd' which is far away from the root node, we have to explore all the nodes from the root node till the depth 'd'.

b) Depth First Search (DFS) is search algorithm in which the strategy of searching is we search all the nodes until we reach the end and get back to first level and then we go for the next node in that level in breadth wise. The search algorithm is stopped when we find the goal node, this type of algorithm is helpful when we have the goal node very far from the starting point. The backlog by using this type of algorithm is if we have the goal node close to the root node, but not in the first few subtrees to be explored then goal node will be found very late as it must travel till end.

c) Iterative Deepening search algorithm is the combination of BFS and DFS it inculcates the space efficiency from DFS and BFS fast search, in this algorithm the node is searched till some depth in breadth first search format, and it is restricted to go beyond that depth. We have to keep increasing the depth until the goal node is found. In this algorithm we have to make a note that the first level nodes are visited each and every time until the goal node is found for each and every depth, whereas second level nodes visited one time less compared to first in trees most of the nodes are at bottom so cost of searching will not effect. Compared to above algorithms Iterative deepening search is most optimized the cost of finding the goal node is very less and the goal node can be found very fast.

d) In Greedy Breadth First Search algorithms[4] among the multiple paths available one path is selected, that path is the local maxima among the paths that are available. Path can also be selected based up on the heuristic function. The heuristic function guides the algorithm to take a particular path which is towards the goal. The heuristic function can be of different types like heuristic value i.e., based up on success rates or number of input nodes.

  i) For each motion success rate is defined, if the success rate is high then the robot can perform that action seamlessly and if the success rate is less then there is high chance that robot may fail. The success rate lies between 0 and 1, if the success rate near to 0 then the robot is not at all familiar with that task. If we have multiple paths which leads to the output path, then we have to select the path based up on the success rate, the one with the high success rate is selected because the robot is familiar with the task and it can do the task seamlessly. For example, if we have two choices for cutting the onion one with knife and one with slicer, the success rate of cutting an onion with slicer is higher than success rate of cutting onion with knife then the robot will select the one with higher success rate i.e., cutting onion with slicer.
  ii) The second type of heuristic function is based up on the number of input nodes, amongst the several paths available the path with the less input nodes is selected. If you are at a point where the same output can be achieved using different paths, one path having the four inputs and the other path with two inputs. The path with two inputs is selected based up on the algorithm.

**Heuristic 1 Algorithm:**

**Input: Goal Node and the ingredients**
**Input: Functional units**
**Input: Items available in kitchen**
**Input: Motion success rates dictionary**
**Input: Empty queue**
**Input: list of items already searched**
**Input: Empty reference task tree**
**Algorithm Maxheuristic**

Add goal node to queue
while queue is not empty do
    l = popping element from the queue
    if l is in items already searched list
        continue
    else
        append l to list to be searched
    c= foon object nodes of l
    if not c in kitchen
    function Maxheuristic:
        for each c
            motion = c.motion_node
            cur[c] = motion
            maxheuval=0
            search_motion = cur[c]
            searchsuccessrate=mot[search_motion]
            cur[c]= search success rate
            calculate max of current motion list
       return maxheuvalue
        end for
selected candidate = call the function
if selected candidate is in reference task tree
continue
append selected candidate to the reference task tree
delete the duplicates if there are any
end if
end while
Output: Reverse the reference task tree

**Heuristic 2 Algorithm:**

**Input: Goal Node and the ingredients**
**Input: Functional units**
**Input: Items available in kitchen**
**Input: Motion success rates dictionary**
**Input: Empty queue**
**Input: list of items already searched**
**Input: Empty reference task tree**
**Algorithm Mininputnodes**

Add goal node to queue
while queue is not empty do
    l = popping element from the queue
    if l is in items already searched list
        continue
    else
        append l to list to be searched
c= foon object nodes of l
if not c in kitchen
create obj_list as empty
function Mininputnodes:
    for each c
        motion = c.motion_node
        object_count = object count of c
        ingred_count=Object.getIngredient(c)
        total_count = object_count+ ingred_count
        append total_count to obj_list
        min value = min(obj_list)
    return min value
    end for
selected candidate = call the function
if selected candidate is in reference task tree
continue
append selected candidate to the reference task tree
delete the duplicates if there are any
end if
end while
Output: Reverse the reference task tree

## IV. Experiments/Discussion

The sub graphs are generated using the algorithms stated below. BFS algorithm will be helpful when the goal node is in the first level and both the heuristic algorithms are greedy search algorithms, the max heuristic algorithm is based up on the success rate value. If we use minimum heuristic algorithms, we can select the path with minimum inputs. Some algorithms will be best based up on the structure and position of the goal node. Based up on the type of algorithm selected to traverse the tree, number of functional units' changes.

TABLE 1: Comparison of number of functional units for different algorithms for five goal nodes.

| Goal Node | BFS | Max Heuristic | Min input Heuristic | Iterative deepening |
|---|---|---|---|---|
| Greek salad | 31 | 31 | 30 | 28 |
| Macaroni | 8 | 8 | 7 | 7 |
| Whipped Cream | 15 | 15 | 14 | 10 |
| Sweet potato | 3 | 3 | 3 | 3 |
| Ice | 1 | 1 | 1 | 1 |

Values in the table 1 illustrates, number of functional units for different algorithms stated above. Functional units for Greek salad dominate among all of them and number of functional units for ice are less. And we can also observe the number of functional units for sweet potato and Ice are same for all the algorithms.

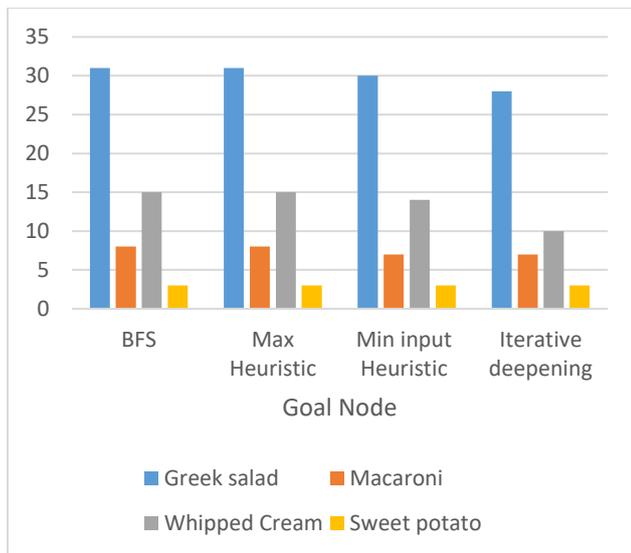

Figure 5: Visualization of data stored in the above table that states the number of functional units for different goal nodes by using different types of algorithms.

As stated in figure 5, that is the graphical representation of functional units with different algorithms. The blue ones represent Greek salad, red one represent macaroni, grey one represents whipped cream and yellow one represents ice.

## V. Conclusion and References

From this paper we can conclude that, robots are very much useful we can do wonders with them. Robots along with human help can lead to next level success in all the sectors. As this paper mainly concentrating on FOON, initially visualization from the video and then generating the task trees and then by using the algorithms we can select the best possible path. After implantation of all the algorithms we have compared the number of functional units for all the algorithms by seeing them we can draw the conclusion that when there are less functional units there is no big difference between all the algorithms but when the nodes are high then there, we can see a difference between the algorithms. Finally, robots along with humans can do wonders.